\documentclass[11pt,a4paper]{article}
\usepackage[hyperref]{acl2019}
\usepackage{amsmath}
\usepackage{times}
\usepackage{latexsym}

\usepackage{url}
\usepackage{graphicx}

\aclfinalcopy 
\def\aclpaperid{1234} 

\newenvironment{myitemize}
{ \begin{itemize}
    \setlength{\itemsep}{0pt}
    \setlength{\parskip}{0pt}
    \setlength{\parsep}{0pt}     }
{ \end{itemize}                  }

\title{Understanding Scanned Receipts}

\author{Eric Melz \\
  300 S Reeves Dr. \\
  Beverly Hills, CA 90212 \\
  \texttt{eric@emelz.com}
}

\date{}

\begin{document}
\maketitle
\begin{abstract}
Tasking machines with understanding receipts can have important
applications such as enabling detailed analytics on purchases,
enforcing expense policies, and inferring patterns of purchase
behavior on large collections of receipts.  In this paper, we focus on
the task of Named Entity Linking (NEL) of 
scanned receipt line items; specifically, the task entails
associating shorthand text from OCR’d receipts with a knowledge base
(KB) of grocery products.  For example, the scanned item ``STO BABY
SPINACH'' should be linked to the catalog item labeled ``Simple Truth
Organic\texttrademark Baby Spinach''.  Experiments that employ a variety of
Information Retrieval techniques in combination with statistical
phrase detection shows promise for effective understanding of scanned
receipt data.

\end{abstract}

\section{Introduction}

Tasking machines with understanding receipts can have important
applications such as enabling detailed analytics on purchases,
enforcing expense policies, and inferring patterns of purchase
behavior on large collections of receipts.  In this paper, we focus on
the task of Named Entity Linking~\cite{Hachey:2012} of
scanned receipt line items; specifically, the task entails
associating shorthand text from OCR’d receipts with a knowledge base
(KB) of grocery products.  For example, the scanned item ``STO BABY
SPINACH'' should be linked to the catalog item labeled ``Simple Truth
Organic\texttrademark Baby Spinach''.  

\section{Related Work}

A literature review reveals virtually no published work in this
specific domain.  While there is a body of work researching text
extraction from scanned receipts (e.g.~\citealp{Huang:2019}), the work is primarily
focused on Named Entity Recognition (NER) instead of Named Entity
Linking (NEL).  That is, systems are considered successful if they can
identify text items such as store locations, totals, etc, but they are
not evaluated with respect to the interpretation of the extracted
text.

Although no papers exist on linking scanned entities, there is
literature in other areas that appear potentially relevant to the
subject task.  This includes work on general-purpose techniques for
building abbreviation dictionaries, acquisition of medical
abbreviations (e.g., ``COPD'' $\rightarrow$ ``Chronic Obstructive Pulmonary
Disorder''), and normalization of social media content (e.g., ``ur
coooool'' $\rightarrow$ ``you are cool'').  The follow sections
summarize a few papers in these areas.

\subsection{Language Independent Acquisition of Abbreviations}
~\cite{DBLP:journals/corr/abs-1709-08074} describe a
language-independent technique for acquiring abbreviations and their
expansions, by exploiting Wikipedia redirect and disambiguation pages.
They begin by motivating the acquisition of abbreviations, noting that
the explosion of social media has made the need for abbreviations
increasingly important.  They also note that a token such as ``ACE''
could have multiple expansions, including ``accumulated cyclone
energy'' and ``American Council on Education'' in addition to the word
``ace'' (as in ``Ace of spades'').

The authors present related work, noting that most of the previous
work for abbreviation detection and expansion extraction has been in
the domain of English biomedical text.  A common strategy is to
identify occurrences where an abbreviation is explicitly paired with
its expansion for example through a pattern involving a parenthetical
such as {\em \verb|<short form>  (<long form>)|} or {\em
  \verb|<long form> (<short form>)|}.
Other approaches consider the contexts of short form and long form
occurrences, pairing short forms with long forms according to their
distributional similarity by measuring the cosine of their context
vectors.  Another approach uses supervised learning, considering
features such as string similarity and other characteristics of the
short and long forms.

The authors' work is based on previous work by~\cite{JACQUET14.468}
who describe a technique for mining abbreviations by making use of
Wikipedia redirection pages.  The authors observe that due to the use
of only redirect pages for the gold standard annotation, a shortcoming
of the prior work is that each abbreviation only has a single
expansion even though multiple different expansions are possible for
some of the abbreviations.  To remedy this shortcoming, the authors
propose mining disambiguation pages in addition to redirect pages to
gather multiple possible long-form expansions.

The authors mine redirect and disambiguation pages for abbreviations,
while applying several rules such as (a) Short forms are restricted to
ten characters or less, (b) At least half of the short-form characters
must be upper case, and (c) The long-form must be at least twice as
long as the short form, with at least two tokens; they generate
candidate expansions and then score the expansions.  Scoring occurs by
computing features for synonym similarity, topic similarity, and
surface similarity.  Synonym similarity means that one term can be
replaced with another while preserving the meaning of the sentence and
is assessed using word embeddings using word2vec~\cite{NIPS2013_5021}.
Topical relatedness means that two terms occur in the same sorts of
documents, and is assessed using Latent Semantic
Analysis~\cite{deerwester-indexing-1990}.  Surface similarity is the
overlap in the surface forms of the terms by computing the best
possible alignment between a short form and a long form.   The three
similarity scores are combined using a logistic regression model.

The authors compare their system with a previous system developed
by~\cite{SchwartzH03} that extracts abbreviations using parentheses
based patterns.  The metric used to compare systems is Area Under the
Precision/Recall curve.  Without the scoring extensions, the 2 systems
are comparable: the Schwartz and Hearst system has an AUC of 0.359 and
the Candidate System has an AUC of 0.324.  However, by adding the
alignment and embedding scoring extensions, the Candidate System’s
performance improves to an AUC of 0.480. 

\subsection{Clinical Abbreviation Expansion}
~\cite{liu-etal-2015-exploiting} describe a system for identifying clinical abbreviation
expansions.  They note that abbreviations are heavily used in medical
literature and documentation.  In notes written by physicians, high
workloads and time pressure intensify the need for using
abbreviations.  This is especially true within intensive care
medicine, where it's crucial that information is expressed in the most
time efficient manner to provide time-sensitive care to critically ill
patients.  Within the arena of medical research, abbreviation
expansion using NLP can enable knowledge discovery and has the
potential to improve quality of care.

The author's system works as follows.  Word embeddings are trained
using word2vec~\cite{NIPS2013_5021}.  The material used to train embeddings consists of
medical  texts such as articles, journals, and books, in addition to
hand-written Intensive Care notes.  To generate expansions for
abbreviations in the hand-written notes, abbreviations are extracted
from the notes, and then matched against a domain-specific
abbreviation knowledge base.  From this list of expansions, 
embedding vectors are retrieved for the abbreviation and candidate
expansion.  A similarity score is computed for each (abbreviation,
expansion) pair, producing a ranked list of candidates expansions.

To test the performance of the system, a ground-truth dataset is
produced by having physicians manually expand and normalize the
handwritten notes.  The authors compare their model against several
baselines.  For example, one baseline chooses the highest rated
candidate expansion in the domain specific knowledge base.  Comparing
accuracy of the author's system against the baselines results in a
50\%+ increase.  For example the rating baseline has an accuracy of 21\%
and the author's system has an accuracy of 83\%.

{Social Media Text Normalization}
~\cite{Lourentzo:2019} present a Sequence to Sequence (Seq2Seq) model
for normalizing social media text.  They observe that social media
texts have an enormous amount of variation, and that text
normalization systems that rely on surface or phonetic representations
may be ill-equipped to handle such variability.  To rectify this
situation, they propose a hybrid word-character Seq2Seq model with
attention.  This type of model has been successfully applied to tasks
such as machine translation, and has promise for text normalization.

The authors frame the task of text normalization as mapping an
out-of-vocabulary (OOV) non-standard word to an in-vocabulary (IV)
standard word that preserves the meaning of the sentence.  The
non-standard forms in user generated content include misspellings   
(defenitely $\rightarrow$ definitely), phonetic substitutions (2morrow $\rightarrow$
tomorrow), shortening (convo $\rightarrow$ conversation), acronyms (idk $\rightarrow$ i don’t
know), slang (``low key'', ``woke''), emphasis (coooool $\rightarrow$ cool), and
punctuation  
(doesnt $\rightarrow$ doesn't).

The authors note that lexicon-based approaches are not able to handle
social media text properly.  String similarity, such as edit distance,
does not work on non-standard words where the number of edits is
large, for example abbreviation.   Additionally, systems that rely on
candidate generation and scoring are limited in that they are not able
to handle multiple normalization errors at once, e.g., spelling errors
on an acronym.  The authors suggest that using end-to-end neural
models, particularly Seq2Seq models can deal with these shortcomings.

The authors train a bidirectional word-based Seq2Seq model to
translate unnormalized text to normalized texts.  OOV words are
trained using a character-based Seq2Seq model.  The dataset is
enhanced by synthetically generating negative examples based on common
normalization transformations.    The network is trained on source
sequences and target sequences.  An example source is ``got exo to
share, u interested?  Concert in hk !'', with a corresponding target of
``got extra to share, are you interested?  Concert in hong kong !''.

The authors present results for several variations of the model,
including a word-level Seq2Seq model and the hybrid word-char Seq2Seq
model.  The best score is an F\textsubscript{1} score of 83.94 on the
hybrid word-char Seq2Seq model.

\section{Data}

\begin{figure}[h]
  \centering
    \includegraphics[width=0.5\textwidth]{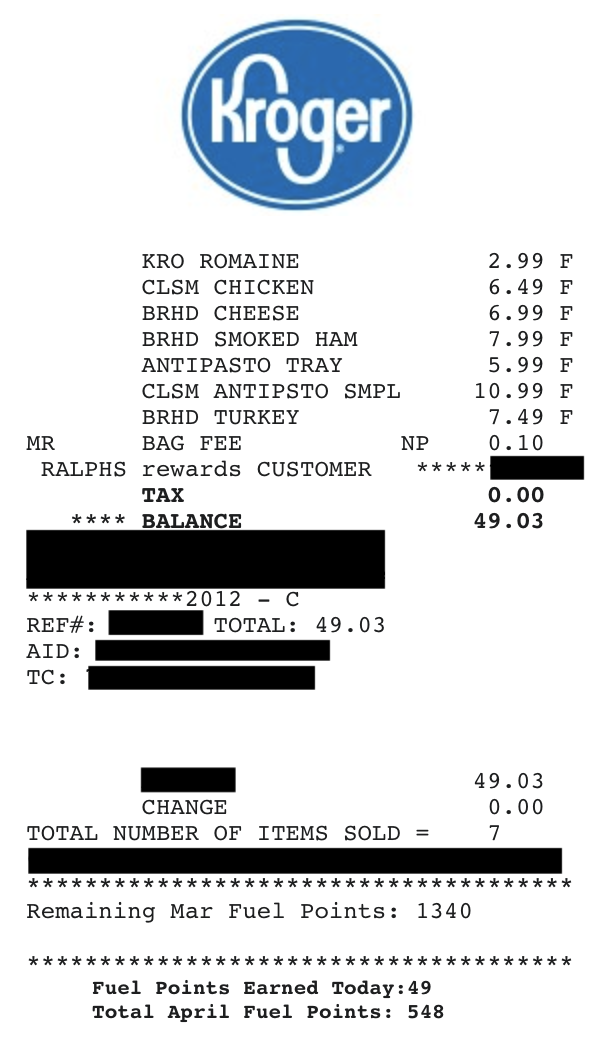}
    \caption{Scanned Receipt}
    \label{fig:scanned}
\end{figure}

For this task, we need a dataset which includes scanned receipt
product mentions (e.g, ``BRHD CHEESE'') and the corresponding product
entities (e.g.,  ``Boar's Head Monterey Jack with Jalapeno Pre-Sliced
Cheese'').  A brief web search revealed that no such publicly available
dataset exists.  To obtain a dataset, we built our own by scraping a
grocery store website that contains purchase data.  Specifically, we
use our personal loyalty account with Ralph's (a subsidiary of
Kroger) to obtain representations of scanned receipts along with
corresponding web pages that contain fully-resolved entities.   
As an example, Figure \ref{fig:scanned} shows an instance of a receipt.

\begin{figure}[h]
  \centering
\includegraphics[width=0.5\textwidth]{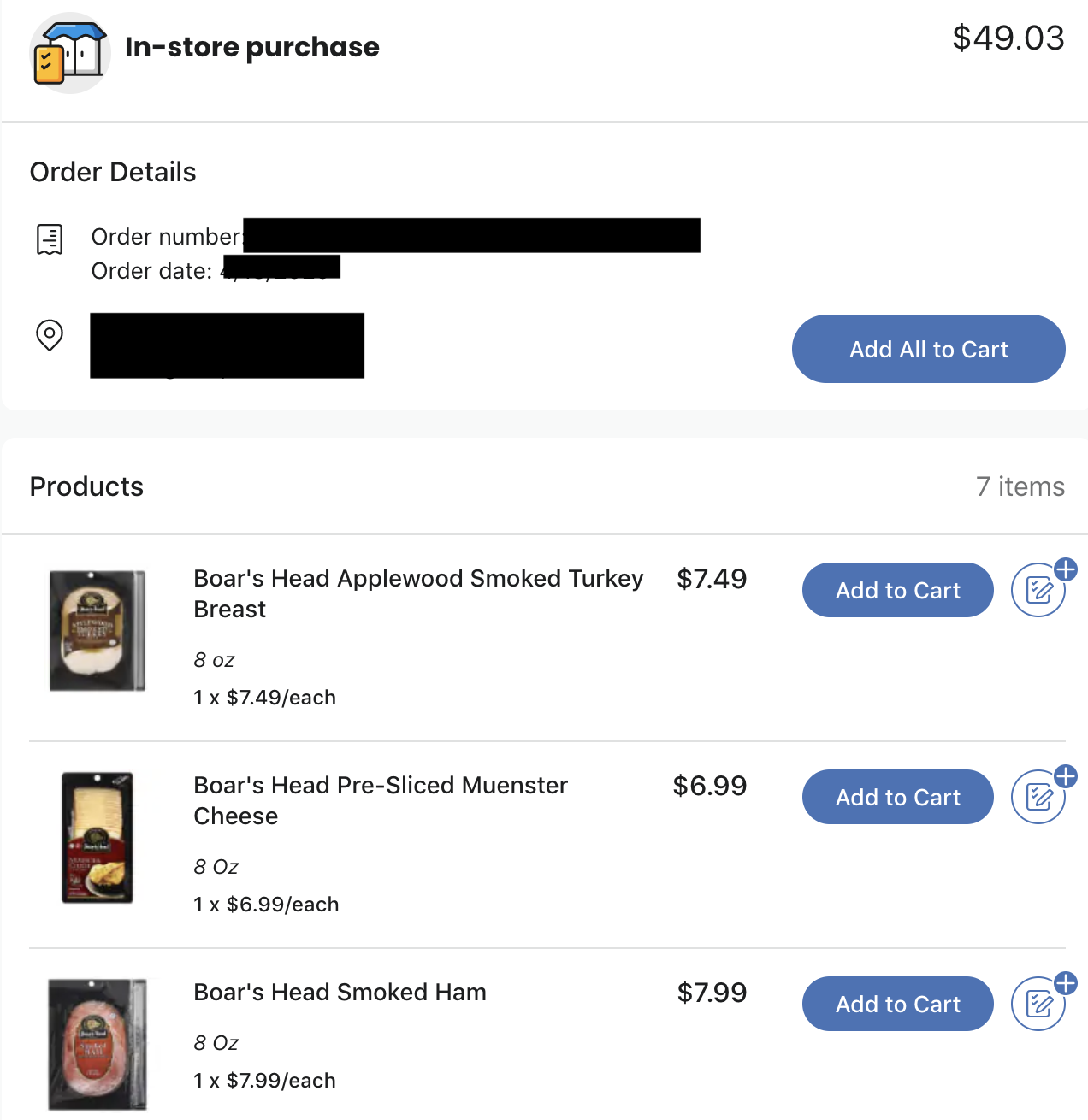}
\caption{Web Receipt}
\label{fig:web}
\end{figure}

Figure \ref{fig:web} shows part of the corresponding web page which contains linked representations of the purchased items.

\begin{figure}[h]
  \centering
\includegraphics[width=0.5\textwidth]{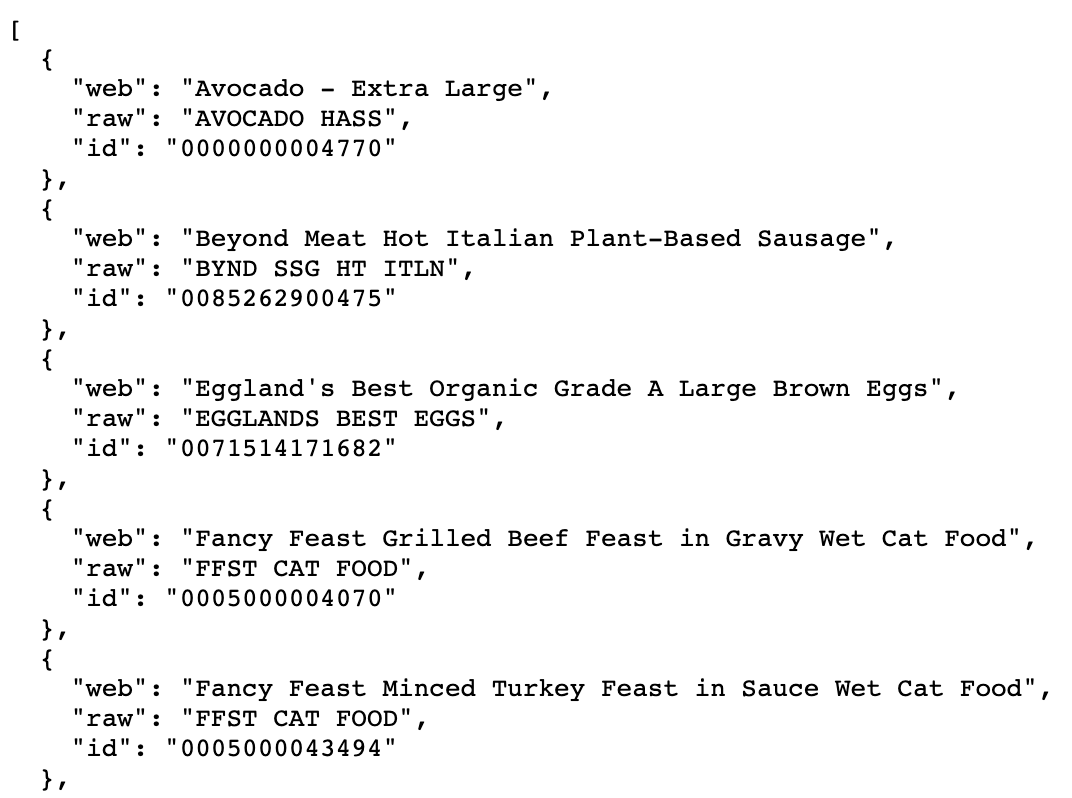}
\caption{JSON representation of joined ``raw'' and ``web'' data}
\label{fig:json}
\end{figure}

We scrape both the text content of the raw receipts and the
user-friendly web rendering, then join the raw receipt data with the
corresponding web data.  This produces a JSON structure per receipt.
A sample of the JSON is shown in Figure \ref{fig:json}.  The ``raw''
field represents the product mention, and the ``web'' field represents
the label associated with the entity.  The ``id'' field is scraped from
the web page and can be used as a succinct identifier for the entity. 

The dataset consists of 65 scraped receipts, producing 711 non-unique
line items, and 296 unique line items.  All data and code for these
experiments are available on Github~\cite{github}.

\section{Methodology}

To evaluate model performance, we gather unique mentions and
measure the accuracy of predicting entities.  This can be conceived as
a multi-class classification task where the entities to be predicted
are the classes.  An alternative metric would be to use a
macro-average F\textsubscript{1} score, but this is overkill for this specific
experiment setup since there is a uniform distribution across classes:
each class is represented by exactly one test instance. 
For example, suppose we have the following two
unique mentions: 
\begin{myitemize}
\item BRHD CHEESE
\item AVOCADO
\end{myitemize}
Further, suppose that the entity for BRHD CHEESE was correctly
predicted as “Boar's Head Monterey Jack with Jalapeno Pre-Sliced
Cheese”, and the prediction for AVOCADO yielded nothing.  The first
prediction is a ``hit'' and the second is a ``miss''.  Dividing the
total hits by the total number of predictions, we obtain an accuracy
of (1 + 0) / 2 = 0.5. 

Note that the mention representations contain
much less information than the entity labels.  In the above example,
BRHD CHEESE is matched with  ``Boar's Head Monterey Jack with Jalapeno
Pre-Sliced Cheese'', but also could have been matched to  ``Boar's
Head Spicy Cheddar Cheese'', or a number of other types of Boar's Head
cheese.  To account for this ambiguity, a prediction is counted as a
“hit” if it is any of the possible resolutions of the product mention.
In the previous example, both of the long descriptions would be
considered hits for BRHD CHEESE. 

The dominant modeling paradigm for the entity linking system we use is
Information Retrieval.   A baseline model indexes entity labels using the Lucene
~\cite{lucene} IR
engine.  Lucene provides a toolkit of tokenizers and token analyzers,
enabling many strategies for matching text.  The most basic setup uses
strict matching on tokens, providing a good baseline model.
Subsequent experiments improve accuracy by selecting more
sophisticated IR and NLP techniques such as using wildcard
queries, phrase detection, etc.

\section{Experiments}

In the following sections, we report a series of experiments that
leverage Lucene.  We index the web versions of entities and query the
index with the raw line items derived from scanned receipts.

Lucene's default scoring formula is
BM25~\cite{robertson:2009}.  BM25 is based on a
bag-of-words approach. The score of a document $D$ given a query $Q$
which contains the words $q_1, ..., q_n$ is given by: 

\begin{equation}
\begin{split}
  \label{eq:bm25}
score(D,Q) = \sum_{i=1}^{n} IDF(q_i) \cdot \\
  \frac{f(q_i, D) \cdot (k_1 + 1)}{f(q_i, D) + k_1 \cdot (1 - b + b
    \cdot \frac{|D|}{avgdl})}
\end{split}
\end{equation}

where $f(q_i, D)$ is $q_i$'s term frequency in the document $D$, $|D|$
is the length of the document $D$ in words,
and $avgdl$ is the average
document length in the text collection from which documents are
drawn.
$IDF$ is the inverse document frequency, calculated as $log(1 + (N - n + 0.5) / (n + 0.5))$, 
where $n$ is the number of documents that $q_i$ appears in, and $N$
is the total number of documents.
$k_1$ and $b$ are free parameters.  $k_1$ represents a term
saturation parameter, controlling the equation's senstivitiy to
incremental increases of term frequencies, and $b$ is a length
normalization parameter, controlling the equation's sensitivity
to the length of a phrase.  For the experiments reported in this
paper, $k_1 = 1.2$ and $b = 0.75$.

To develop and optimize the retrieval process, we worked with an index
derived from a single receipt.
In each section, we report scores against an index based on the
development receipt.  The final results report results against
the entire set of 65 receipts.

\subsection{Baseline}

The first experiment used out-of-the-box Lucene.  This setup tokenized
the index and queries using the $StandardAnalyzer$, which does
superficial processing on terms such as lowercasing them.

Using this approach, we acheive an accuracy of 0.63.
This basic approach is capable of matching entities with identical
scanned representations, but falls short when the scanned
representation contains abbreviations.  For example, the
query ``KRO WATER'' will consider the entity ``Fiji Water''
an equally good match as ``Kroger Water'' even though ``KRO''
is an abbreviation for ``Kroger''.

\subsection{Wildcards}

To help solve the problem of missed abbreviations, we introduced a
wildcard technique.  First, we construct a dictionary of all the words
present in the web entity entries.  When constructing a query from the
scanned representation, we eliminate all words that are present in the
dictionary.  For the remaining terms, we rewrite those terms to match
indexed terms that contain all the letters in the raw terms.  For
example, the term ``KRO'' will be rewritten as in the query as
``K*R*O*'' and hence match ``Kroger''.  Note that in Lucene, wildcard
matches do not use term frequency or inverse document frequency but
count as 1.0 if there is a match between the wildcard and some indexed
term, and 0 otherwise.  Hence, a query with 3 wildcard terms and two
matches would receive a score of exactly 2.0.

Using the wildcard technique, we boost the accuracy to 0.84, a
substantial improvement of the baseline of 0.63.  The wildcard
approach does seem to make some progress towards solving the
abbreviation problem, however, there are still cases where this simple
approach fails.  For example, if an abbreviation spans multiple terms,
the wildcard approach will not be able to find a match.  For example,
the query ``P*R*S*L* TOMATOES'' will not match the entity ``Private
Select Tomatoes'', since wildcards are only matched against single
terms: ``P*R*S*L*'' matches neither ``Private'' or ``Select''.

\subsection{Mashed Wildcards}

To attempt to solve the multi-word abbreviation problem, we introduced
a new field into the indexed documents, called ``mashed\_terms''.  This
field concatenates all terms into a set of mashed
terms, allowing each word in the set to serve as the prefix of the
mashed term.  For example, the terms ``the quick brown'' would
be rewritten into the mashed\_terms field as ``thequickbrown quickbrown
brown''.  This approach matches multi-word abbreviations.  For example
``P*R*S*L*'' will now match ``privateselect''.  Using the Mashed
Wildcards approach, accuracy improves to .88. 

\subsection{Phrases}

Examination of errors using the Mashed Wildcard technique reveals that
mashing terms results in some strange false positives.
For example,``OCEANS HALO BROTH'' matches ``Pero Organic Green Beans''
This is due to the fact that the non-dictionary term ``OCEANS''
matches the mashed term ``OrganiCgreenbEANS '' (caps indicate matching
characters).  This type of error suggests that we should be more
selective when generating mashed terms.  Ideally, we would like to
curate semanticially meaningful phrases instead of long, arbitrary
sequences of words.  To do this, we can analyze the entity labels and
identify strongly collocated bigrams and trigrams, and use
concatenated versions of those instead of mashed terms.

An information-theoretically motivated measure for discovering
intersting collocations is {\em Pointwise Mutual Information}~\cite{manning:1999}

\begin{equation}
\begin{split}
  \label{eq:pmi}
  I(x^\prime,y^\prime) = log_2\frac{P(x^\prime,y^\prime)}{ P(x^\prime)P(y^\prime) } \\
  = log_2\frac{P(x^\prime|y^\prime)}{ P(x^\prime) } \\
  = log_2\frac{P(y^\prime|x^\prime)}{ P(y^\prime) } \\
\end{split}
\end{equation}

Here, $x^\prime$ and $y^\prime$ represents two terms of interest.
Using this technique, we generate a list of semantically significant
bigrams and trigrams.  For example, the top 10 detected bigrams are:
\begin{myitemize}
\item advanced whitening
\item alfaros artesano
\item alfresco pasture-raised
\item ang chck
\item antipasto italiano
\item arm hammer
\item arrabbiata fra
\item aretsano bakery
\item athenos crumbled
\item atkins endulge
\end{myitemize}
And the top 10 trigrams are
\begin{myitemize}
\item alfaros artesano bakery
\item ang chck ptty
\item antipasto italiano wildbrine
\item arm hammer peroxi
\item arrabbiata fra diavolo
\item artesano bakery bun
\item atkins endulge chocolate
\item bagel sesame bagels
\item bagels zia italiana
\item bf ang chck
\end{myitemize}
Using the ngram technique, we improve accuracy to .90

\subsection{Fuzzy Phrases}
It turns out that abbreviations are not the only match problem that we
need to contend with.  Sometimes, raw representations are represented
as plural, whereas the web representations are represented as
singluar.  For example, the web representation could be ``artichoke''
with a corresponding raw representation of ``ARTICHOKES''.  A query
of ``A*R*T*I*C*H*O*K*E*S*'' will not match ``artichoke''.  To deal
with this, we introduce fuzzy matching, which will match terms within
a small edit distance of the query term.  In the above example, the
query ``ARTICHOKE'' will be rewritten to ``A*R*T*I*C*H*O*K*E*S*
ARTICHOKES\textasciitilde''.  The second term will correctly match the ``artichoke''
entity.

Using fuzzy terms in conjunction with ngrams, accuracy improves to
0.93. 

\section{Results}

We ran each IR method on an index of all 65 receipts.
Table \ref{table:results} shows the accuracies for the single receipt and
all receipt cases.  The results show improvement with the introduction of
each new IR technique.  Unsuprisingly, the results for all receipts is
lower than that of a single receipt.  This is primarily due to the
expansion of choices introduced by more data.  Recall that we are
only judging a hit by looking at the top search result.  It is likely
that in many examples, the true match is still contained in one of the
top-k matches, where k is a relatively small number.

\begin{table}[t!]
\begin{center}
\begin{tabular}{|l|r|r|}
\hline \textbf{Technique} & \textbf{Single Receipt} & \textbf{All Receipts} \\ \hline
Baseline & 0.62 & 0.47 \\
Wildcard & 0.84 & 0.72 \\
Mashed & 0.88 & 0.76 \\
Ngrams & 0.89 & 0.77 \\
Fuzzy Ngrams & 0.93 & 0.79 \\
\hline
\end{tabular}
\end{center}
\caption{\label{table:results} Receipt Item Linking Results. }
\end{table}

\section{Synonym Expansion}
We explored the possibility that synonym expansion of queries
could produce enhanced accuracies.  The intuition is that by
generating synonyms of the raw forms, recall of the web forms
could be improved.  For example a synonym of ``avocado'' is
``avocados'' and such an expansion could improve accuracy without
using fuzzy matching.  Similarly, a closely related word to
``cheddar'' is ``cheese'' and such a synonym expansion could be
potentially beneficial.

We explored the possibility of synonym expansion using
word2vec~\cite{NIPS2013_5021}.  First, we trained word2vec on our web
phrases and informally tested it by generating similar words for
various words of interest.  The results of these experiments were
not promising.  For example, generating similar words for
``cheese'' produces
\begin{myitemize}
\item feast
\item french
\item smoked
\item beyond
\item grade
\item ground
\item wet
\item hot
\end{myitemize}  

The results indicate that using word2vec with our current receipt
dataset will not produce very good synonyms.  This could be due
to the small size of the dataset and the limited context that
entity descriptions provide.

Another thing we tried is training word2vec on the Brown
Corpus~\cite{francis79browncorpus}, and anectodally examining
similarity.  This produced similarly poor results.  For example,
querying on ``cheese/nn'' produces
\begin{myitemize}
\item lime/nn
\item editors/nn
\item vivid/jj
\item jokes/nn
\item grapes/nn
\item uneven/jj
\item corn/nn
\item loose/jj
\item brutal/jj
\end{myitemize}  

Although it's possible that filtering results to only nouns could
produce better results, the fact that ``editors'', ``grapes'' and
``corn'' are considered related to ``cheese'' does not inspire a lot
of hope for this approach.

\section{Future Work}

Our experiments show that Named Entity Linking (NEL) on scanned
receipts can be effectively tackled by employing a variety of
off-the-shelf Information Retrieval (IR) techniques.  Using
successive optimizations in our approach, we improved accuracy from a
baseline of 0.47 to 0.79, a 68\% improvement.

One thing to note about our approach is that it is heavily reliant
on lexical features of the text and pays little attention to the
actual semantics of the terms in our domain.  Future explorations
should focus on techniques for effective synonym expansions,
possibly using resources such as WordNet~\cite{wordnet} or other
domain-specific lexicons.  It is also possible that training
word2vec with a sufficiently large receipt dataset or related
resources could produce reasonably accurate similarity dictionaries.

Note that some matches are particularly challenging and would
require a deep knowledge about the receipt domain.  For example,
the surface form ``CA REDEM VAL'' should be matched to the entity 
``CRV DEPOSIT''.  While there is no surface similarity between these
two expressions, they both refer to recycling deposits.  A deeper
analysis of terms, phrases and their meanings could prove beneficial
in cases such as this.

Another possibility is that neural Seq2Seq models could be used to
``rewrite'' raw queries into their final entity forms.  This problem
resembles machine translation.  The main impediment to this approach
is the amount of data that is need to produce quality results.

\section{Conclusion}
Named Entity Linking of scanned receipt data is an important problem
with many potential applications.  Using a variety of IR techniques,
we are able to obtain a 68\% improvement in accuracy over a baseline
system.   Future work includes exploring the possibility of leveraging
neural models to improve semantic matching capabilities of the system.

\bibliography{receipt_nlu}
\bibliographystyle{acl_natbib}

\end{document}